\documentclass[conference]{IEEEtran}
\IEEEoverridecommandlockouts
\usepackage{cite}
\usepackage{amsmath,amssymb,amsfonts}
\usepackage{algorithmic}
\usepackage{graphicx}
\usepackage{textcomp}
\usepackage{xcolor}
\usepackage[ruled,vlined]{algorithm2e}
\usepackage{verbatim}
\def\BibTeX{{\rm B\kern-.05em{\sc i\kern-.025em b}\kern-.08em
    T\kern-.1667em\lower.7ex\hbox{E}\kern-.125emX}}
    
\newcommand*{\myfont}{\fontfamily{<qzc TeX Gyre Chorus>}\selectfont}
\DeclareTextFontCommand{\textmyfont}{\myfont}
\begin{document}

\title{Assessing and Accelerating Coverage in Deep Reinforcement Learning}

\author{\IEEEauthorblockN{Arpan Kusari}
\IEEEauthorblockA{\textit{University of Michigan Transportation Research Institute} \\
\textit{University of Michigan}\\
Ann Arbor, Michigan, USA \\
kusari@umich.edu}
}

\maketitle

\begin{abstract}
Current deep reinforcement learning (DRL) algorithms utilize randomness in simulation environments to assume complete coverage in the state space. However, particularly in high dimensions, relying on randomness may lead to gaps in coverage of the trained DRL neural network model, which in turn may lead to drastic and often fatal real-world situations. To the best of the author's knowledge, the assessment of coverage for DRL is lacking in current research literature. Therefore, in this paper, a novel measure, Approximate Pseudo-Coverage (APC), is proposed for assessing the coverage in DRL applications. We propose to calculate APC by projecting the high dimensional state space on to a lower dimensional manifold and quantifying the occupied space. Furthermore, we utilize an exploration-exploitation strategy for coverage maximization using Rapidly-Exploring Random Tree (RRT). The efficacy of the assessment and the acceleration of coverage is demonstrated on standard tasks such as Cartpole, highway-env.
\end{abstract}


\section{Introduction}
Sequential decision making is an essential component of robotics, utilizing a sequence of actions (chosen from a set of possible actions) to navigate an uncertain environment to best achieve a goal. Reinforcement learning \cite{sutton2018reinforcement} has been proposed to provide a formal approach to solve the sequential decision making problem where an agent chooses actions so as to maximize the cumulative rewards. Over the last decade, deep reinforcement learning (DRL), which combines RL with deep neural networks, has been applied successfully to to various applications with continuous state space and discrete/continuous action space. The wide-ranging applications where DRL has been successfully applied include board games (e.g., Go \cite{silver2017mastering} and Chess \cite{silver2018general}), video games (e.g., Atari games \cite{mnih2013playing}), and complex robotics control tasks (e.g. \cite{levine2016end}).  

Currently DRL tasks utilize reward per episode as the common metric for measuring the performance of the model and completion of the training process. However, there is an inherent assumption placed on the simulation, which generates samples for the training, to be able to sample uniformly from the entire sample space and provide adequate coverage at the end of the training process. Coverage has been defined as the measure of how much of the Design and Verification Environment (DVE) has been covered via testing\cite{hollander_2016}. There is extensive literature on coverage driven verification (CDV) for functional verification of hardware designs \cite{araiza2015coverage}.  Coverage testing for assurance of machine learning models has been explored by transforming high-dimensional data into a lower dimensional manifold in \cite{byun2020manifold}. CDV has also been applied successfully to Deep Neural Networks (DNN) by developing four different dataset quality measures and measuring the quality of most common test datasets on popular DNN models \cite{mani2019coverage}. However, to the best of the author's knowledge, there has been no research conducted exploring coverage testing in the context of DRL. Assessment of coverage appears to be an important factor which can influence the acceptance of the DRL model in a real-world setting. 

 Coverage testing in traditional development relies on functional requirements or model-based techniques, which are, by their very nature, deterministic and provides a tractable solution. Assessment of coverage is complicated in DRL for three primary reasons: 
\begin{itemize}
    \item Data for training a DRL is generated by the simulation environment during the training process which can make coverage testing difficult
    \item There is no explicit model (or functional requirements) in model-free DRL to rely on in order to do exhaustive coverage testing
    \item The DRL tasks are often characterized by a high dimensional continuous state space which suffers from the "curse of dimensionality" \cite{bellman1966dynamic} and makes calculation of coverage problematic
\end{itemize}

Since DRL tasks involve high-dimensional state space, reducing the dimensionality of the state space by projecting the state samples to a lower dimensional manifold makes the computation of coverage tractable. In this paper, we reduce the state space to two dimensions using t-distributed Stochastic Neighbor Embedding (t-SNE) \cite{maaten2008visualizing}. Thus, we get a pseudo measure of coverage where more is inherently better without equating it to actual coverage. Even after reducing the dimensionality to two, the simulation environment generates a lot of samples during the training process and computing the exact area of these samples is often a complex problem. Therefore, we implement an over-approximation of the coverage by utilizing an occupancy grid-like approach \cite{moravec1985high}. Using a fine grained 2-D grid, we apply a voting process of whether a particular point falls within a grid cell. We then calculate the ratio of number of occupied grid-cells (non-zero grids) to the total number of grid-cells. We refer this measure to Approximate Pseudo-Coverage (APC).   

For coverage maximization, we note the following two observations: 
\begin{itemize}
    \item Currently, the simulation environments for DRL are blind to the training process and thus, are not adaptive with the changes in the training process. Our hypothesis is that providing a feedback to the simulation environment would help in maximizing the coverage.
    \item The starting state has a large impact on the evolution of the episode \cite{tavakoli2018prioritizing}. Searching for unexplored areas of the state space can provide a maximization in coverage. Also, small perturbations of high scoring start states can help us in maximizing rewards as well.
\end{itemize}
With these observations in mind, we introduce an exploration-exploitation strategy for determining the start state for each episode. Using an epsilon-greedy policy, we explore using a Rapidly Exploring Random Tree (RRT) \cite{lavalle1998rapidly} to efficiently search high-dimensional non-convex spaces and exploit using a small random perturbation of highest scoring start state. 

The paper is organized as follows: Section \ref{sec:background} provides a background on deep RL and coverage testing, Section \ref{sec:methodology} details the coverage assessment and maximization procedure, Section \ref{sec:results} presents the results of the proposed procedure on standard tasks and Section \ref{sec:conclusions} gives the discussions and conclusions.

\section{Background}
\label{sec:background}
\subsection{Coverage testing}
Coverage is defined as a measure of completeness of set of tests for checking a model of a design. In coverage based verification, the planning of coverage is performed before the start of the verification process \cite{piziali2008coverage}. Each measurable task is known as a coverage task and together, they form a coverage model \cite{benjamin1999study}. Pseudorandom generation techniques are used in test generators in CDV \cite{araiza2015coverage}. CDV uses functional requirements or models to create this test suite. In model-based test generation, a model is explored or traversed to obtain abstract tests, i.e. tests at the same level of abstraction as the model \cite{lackner2012modeling}.

There are different kinds of coverage which can be broadly classified into two categories, normal coverage and pseudo-coverage \cite{hollander_2016}. These constitute:
\begin{itemize}
    \item Normal coverage - This refers to exact coverage testing where achieving 100\% explains the completeness of the tests for DVE. There are two kinds of coverage testing which fall in this category:
    \begin{itemize}
        \item Implementation coverage - measuring which lines of code have been executed or how many state bits have been toggled
        \item Functional coverage - user defined metric intended to investigate to which extent the functionality of a given design under test (DUT) has been verified
    \end{itemize}
    \item Pseudo-coverage - This type of coverage testing relies on measures where more coverage implies better without having a exact upper limit. Examples include:
    \begin{itemize}
        \item Novelty coverage - measures if a new sample is sufficiently far from the current samples
        \item Criticality coverage - measures the closeness of the samples from the terminal state
    \end{itemize}
\end{itemize}
Coverage in the context of machine learning models is not a thoroughly studied problem. With deep learning algorithms being increasingly used in safety critical applications, assessing and accelerating coverage for such applications is of paramount importance. 

\subsection{Deep Reinforcement Learning}
The central idea behind reinforcement learning (RL) is to learn the sequence of actions that an agent takes to maximize the cumulative reward. In an RL task, for any given state $s_t \in S$, the exact action is not properly defined a priori and the agent needs to randomly choose actions $a_t$ from a set of actions, $A$, to find the optimal action at that state based on the q-value of action $Q(s_t, a_t)$. The mathematical formulation governing this process is known as Markov Decision Process (MDP), where the next state is completely defined by the current state and action (Markov property) and choice of an action leads to a scalar reward \cite{sutton2018reinforcement}. 

The RL problem is exactly defined and solved for a discrete state space. However, for a continuous state-space problem, function approximators have to be utilized to learn the value function and/or policy function. Deep neural networks being universal function approximators, can be used to learn such functions. The problem that appears then is that exact convergence is not known and thus, practitioners have to resort to using reward per episode as the common metric to assess if the training is complete. 

\section{Methodology}
\label{sec:methodology}
\subsection{Coverage assessment}
The goal of coverage assessment is to find the gaps in coverage of test data for the model under test. The pseudo-code for the coverage assessment algorithm is given in Algorithm \ref{algo:coverage_assessment}. The first issue in getting a true coverage estimate is that the state space in DRL applications is inherently high-dimensional. For an efficient computation of coverage, we need to reduce the dimensionality of the samples from the simulation. The preliminary requirement in such a dimensionality reduction is that the local neighborhood be preserved i.e. similar points in the high dimensional space should appear close by in the low-dimensional embedding. In this paper, we utilize t-SNE, a nonlinear dimensionality reduction technique proposed by \cite{maaten2008visualizing} which embeds data from a high-dimensional setting onto a two or three dimensional space. t-SNE constructs a probability distribution over pairs of high-dimensional samples such that neighboring points are assigned higher probability while distant points are assigned lower probability. Then, a low-dimensional embedding is chosen such that the Kullback-Leibler divergence (KL divergence) is minimized with the high-dimensional probability distribution.

The other issue in developing a coverage assessment technique for DRL is that the data is generated using a simulation concurrently with the training process. The extents of the coverage are, thus, unknown a priori. We create a buffer for storing the state samples and after a certain `k' number of episodes, create a two-dimensional t-SNE embedding from the samples. The pseudo-coverage for the training process until the current episode can then be defined as the area covered by these embedded state samples. There are different approximations for calculation of area of scatter samples. A common one is estimating convex hulls of the sampled points. However, for points not in a single cluster, the calculation of convex hulls is complex and not completely necessary. In this paper, we utilize an over-approximation using a regularly spaced 2-D grid. The idea of the evenly spaced grid is similar to occupancy grid mapping \cite{moravec1985high} where the non-zero grid-cells represent the coverage of the embedded states. Based on the current extent of the embedded states, we create a grid with a user-defined cell size. The cell size is an important parameter, where there is a trade-off between granularity and computation. The voting process calculates the histogram of the 2-D points and returns the non-zero grid-cells. The buffer is reset and the process is repeated for the next `k' episodes with the non-zero grid-cells being added to the previous count. APC is calculated as the ratio of non-zero grid-cells to the total number of grid-cells. 

\begin{algorithm}
\label{algo:coverage_assessment}
\SetAlgoLined
\KwResult{Approximate Pseudo-Coverage }
$grids \gets Empty$\;
\While{training} {
    $statesBuffer \gets Empty$\;
    Run k episodes and collect statesBuffer\;
    $statesEmbedded \gets RunTsne(states buffer)$\;
    $limits \gets CalculateLimits(states embedded)$\;
    \eIf {$grids == Empty$} {
        $grids \gets CreateGrid(limits, cell size)$\;
    }{
        $grids \gets AdjustGrid(grids)$\;
    }
    $voteGrids \gets Histogram(statesEmbedded, grids)$\;
}
$APC = \dfrac{count(NonzeroGrids)}{count(grids)}$\;
\caption{Coverage assessment}
\end{algorithm}

\subsection{Coverage maximization}
While the process of executing more and more episodes would continue to increase the coverage asymptotically, it might take a long time to fill the gaps in coverage, esp. for high-dimensional state space problems. Coverage maximization automates this process leading to faster coverage attainment. The idea behind maximizing coverage is to generate a feedback loop to the simulation, creating an adaptive process. We detail the procedure for coverage maximization in Algorithm \ref{algo:coverage_maximization}. In order to balance exploration of the unknown initial states with the exploitation of known high-scoring initial states, we implement an epsilon-greedy policy for determination of initial states. 

In the beginning of the training process, we want to find unique initial states which are in the neighborhood of the known initial states. We implement a RRT structure to find these unique initial states. RRT is designed to efficiently search non-convex, high-dimensional spaces such that the tree is constructed by biasing the exploration towards unexplored portions of the space \cite{lavalle2001rapidly}. The idea is to cover the entire state space efficiently without any coverage gaps. 

For the exploitation phase, we want to explore the vicinity of high-scoring initial states. We utilize an idea from neuroevolution \cite{such2017deep} where the best performing agent (described as parent) is mutated by applying a small Gaussian noise to the initial state. Therefore, we pick the highest scoring initial state (or select randomly from equally highest scoring states) and perturb the state by a small amount. This provides adequate coverage around the potentially high-scoring initial states. 

\begin{algorithm}
\label{algo:coverage_maximization}
\SetAlgoLined
\KwResult{initStatesList}
initialize $epsilon \gets 1$\;
initialize $epsilonDecay \gets 0.998$\;
initialize $rrt \gets beginNode$\;
initialize $rrtDist$\;
initialize $initStateList \gets Empty$\;
\While{training} {
    \eIf{$random \leq epsilon$}{
        $initStateCandidate \gets Random(stateSize)$\;
        $neighborNode \gets GetNeighborNode(initStateCandidate)$\;
        $newNode \gets CreateNode(neighborNode, rrtDist)$\;
        $ rrt \gets Append(newNode)$\;
    }
    {
        $maxInitState \gets Max(initStatesList)$\;
        $initState \gets Perturb(maxInitState)$\;
    }
    $initStatesList \gets  Append(initState)$\;
    $epsilon \gets epsilon * epsilonDecay$\;
}
\caption{Coverage Maximization}
\end{algorithm}

\section{Results}
\label{sec:results}
The coverage assessment and maximization technique is validated in two example environments: Cartpole \cite{brockman2016openai} and highway driving \cite{highway-env}. 

\begin{figure*}[h!]
    \centering
\includegraphics[scale=0.08]{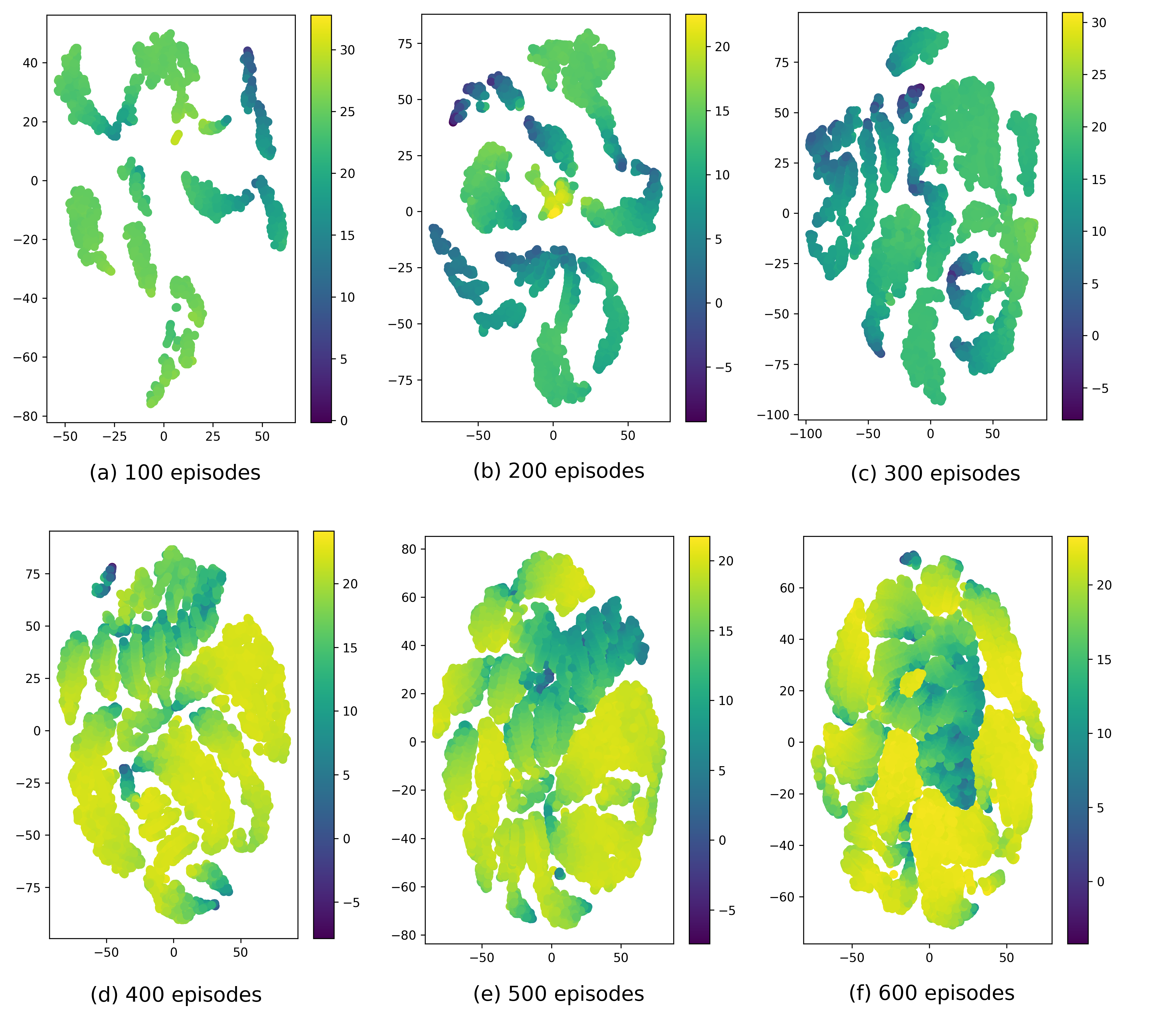}
         \caption{Visualizations of embedded states colored by the q-value of that state with the first action at every 100 episodes. }
        \label{fig:ori_states}
\end{figure*}

\begin{figure*}[h!]
    \centering
\includegraphics[scale=0.08]{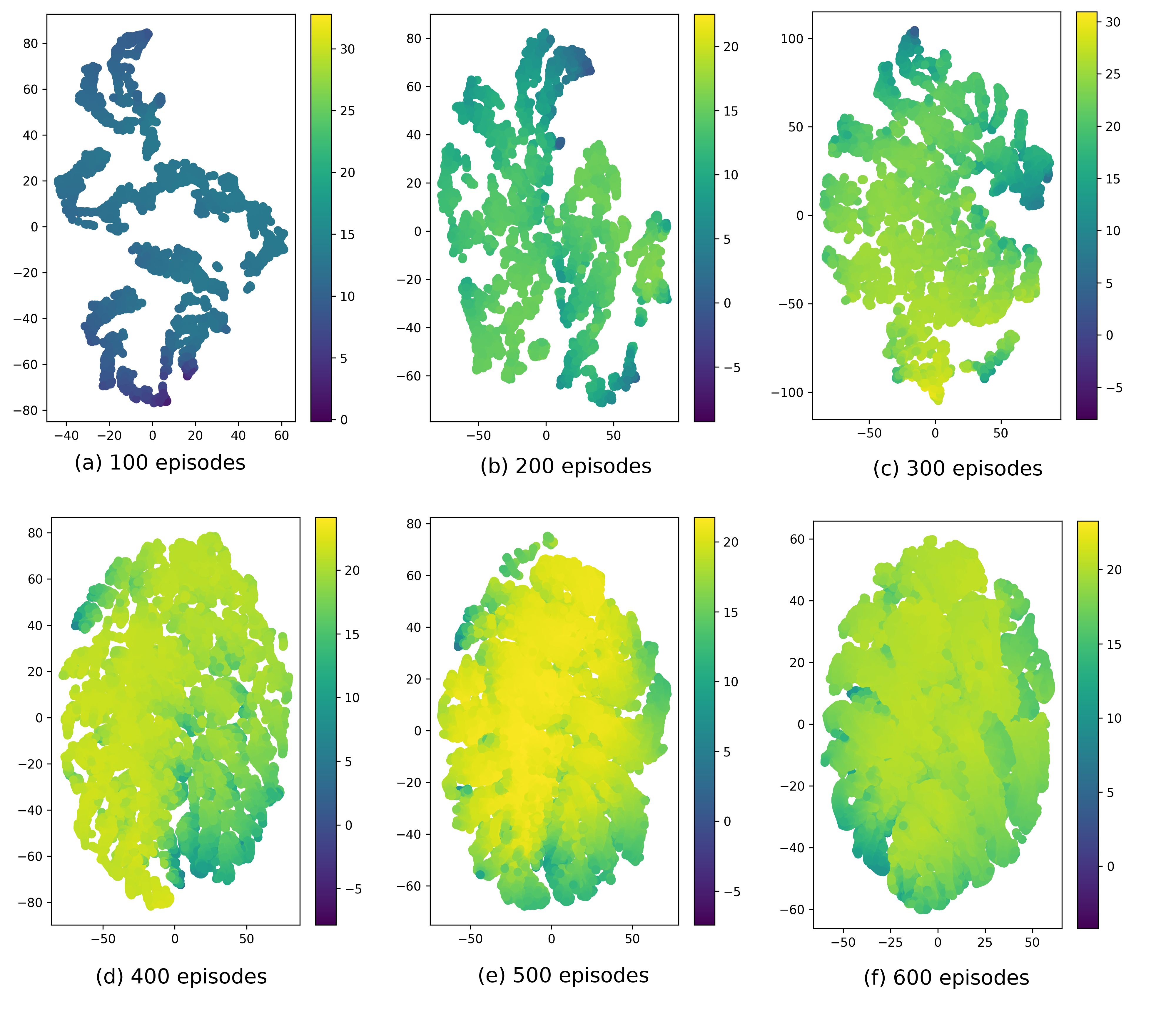}
         \caption{Visualizations of embedded states derived using coverage maximization colored by the q-value of that state with the first action at every 100 episodes. }
        \label{fig:rrt_states}
\end{figure*}

\begin{figure*}[h]
    \centering
\centerline{\includegraphics[scale=0.16]{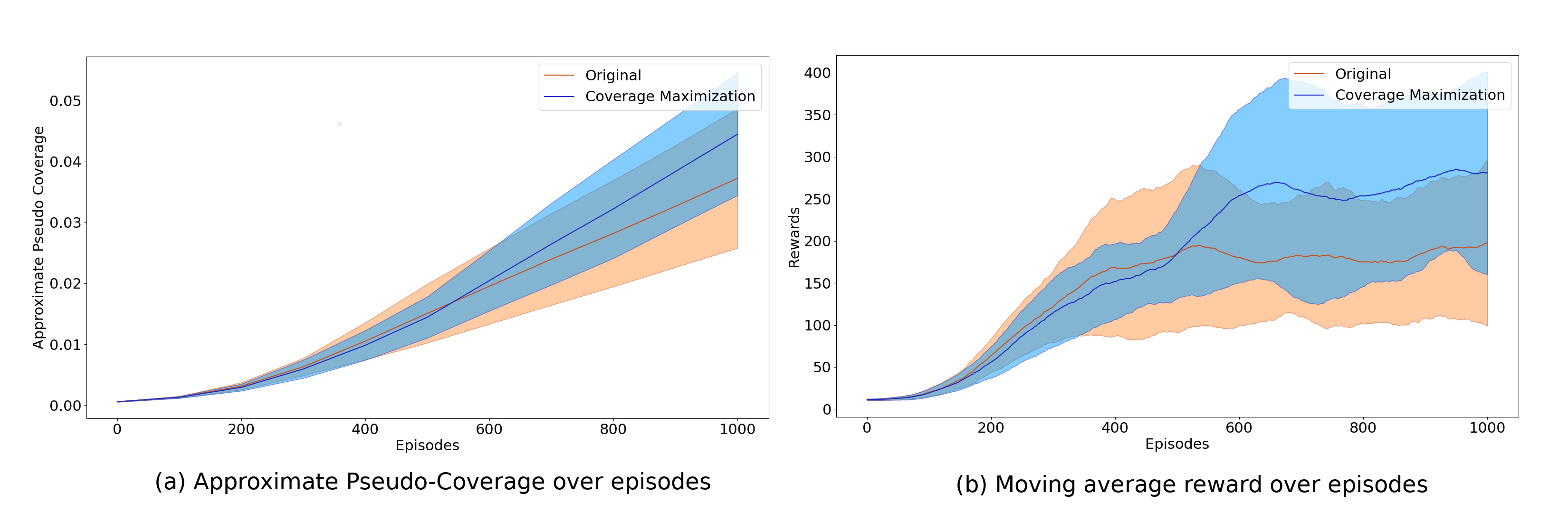}}
         \caption{Comparison of mean and bounds of (a) pseudo-coverage metrics and (b) reward per episode for original method and our proposed method with coverage maximization strategy for 50 runs}
        \label{fig:comparison_cartpole}
\end{figure*}

\subsection{Cartpole}
The cartpole environment \cite{brockman2016openai}, also known as the inverted pendulum problem, is a well-known problem in RL whereby a pole is attached to a cart. The pole is unstable and the only way to keep it upright is by moving the cart horizontally left or right by one unit. The episode ends when the pole is more than 15$^\circ$ from vertical, or the cart moves more than $2.4$ units from the center. The state vector is composed of the position of the cartpole, the angle of the cartpole, and their derivatives. The default reward is the amount of time the pole is upright in a given episode (time limit of 500 timesteps). Due to the nature of the continuous state space, it is impossible to visit every state and action pair during training or to know when the training is complete. OpenAI judges the training to be complete when the agent gets an average of 475 (out of a possible 500) or above for 100 consecutive episodes. 

Deep Q-network (DQN) has been proposed by \cite{mnih2015human} which provided the first algorithm which stably combines deep neural networks with RL to solve continuous state discrete action problems. DQN uses a neural network that gives the Q-values for every action and uses a buffer to store old states and actions to sample from which helps to stabilize training. The cartpole environment is solved using the DQN approach.

In order to provide an intuition of the pseudo-coverage for the state samples, we plot the embedded state samples colored by the q-value corresponding to the first action at every 100 episodes in Figure \ref{fig:ori_states}. The figure shows that even at 600 episodes, there are gaps in the embedded state space which leads to gaps in coverage. Also, we find that in the initial episodes, the state samples form sparse non-convex clusters finding the area of which is a costly and time-consuming problem. The over-approximation using a grid provides an inexpensive solution while keeping over-approximation bounded.

Correspondingly, we utilize coverage maximization to determine the initial state samples keeping the random seed constant for both implementations. We use the position and orientation of the cartpole as an abstraction of the model in the RRT structure. This makes the computations efficient while still allowing for sufficient exploration of the state space. The embedded state samples for the environment with coverage maximization is shown in Figure \ref{fig:rrt_states}. Visually, we can see that the space utilization using coverage maximization strategy is higher than normal starting from the early episodes and by 400 episode mark, there are almost no gaps present in the embedded state samples. This provides a visual confirmation that coverage maximization does bias the system to fill gaps in the coverage. To get some statistics on the efficacy of the coverage, we run the original implementation and the proposed implementation 50 times. The mean and standard deviation of the APC and the reward per episode is given in Figure \ref{fig:comparison_cartpole}. We observe that the proposed method is effective in increasing the coverage by 19.5\% and the reward per episode by 42.3\% at the end of 1000 episodes.  

\subsection{Highway driving}
\begin{figure}[h]
    \centering
\centerline{\includegraphics[scale=0.25]{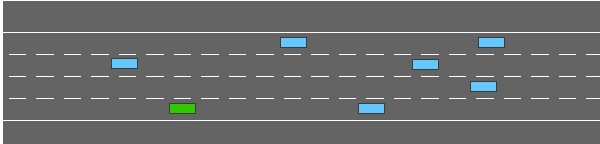}}
         \caption{Highway driving simulation environment}
        \label{fig:highway_env}
\end{figure}

We implement an environment for simulated highway driving and tactical decision making provided by \cite{highway-env}. We choose this environment as a representative example of high dimensional environment with a lot of variability. The environment is built on top of OpenAI gym environment and features an easy-to-use abstraction layer for training DRL to navigate an autonomous vehicle through traffic. Figure \ref{fig:highway_env} shows the layout of the simulation environment. The green rectangle denotes the ego-vehicle while the blue rectangles denote the other vehicles. The obstacle vehicles are capable of changing lanes with each vehicle making independent lane-change decisions at every instant. The ego-vehicle motion is described by kinematic bicycle model\cite{polack2017kinematic}. The longitudinal motion of the other vehicles are governed by the Intelligent Driver model \cite{treiber2000congested} while the lane change decisions are governed by the MOBIL model \cite{kesting2007general}. 

\begin{figure*}[h]
    \centering
\centerline{\includegraphics[scale=0.16]{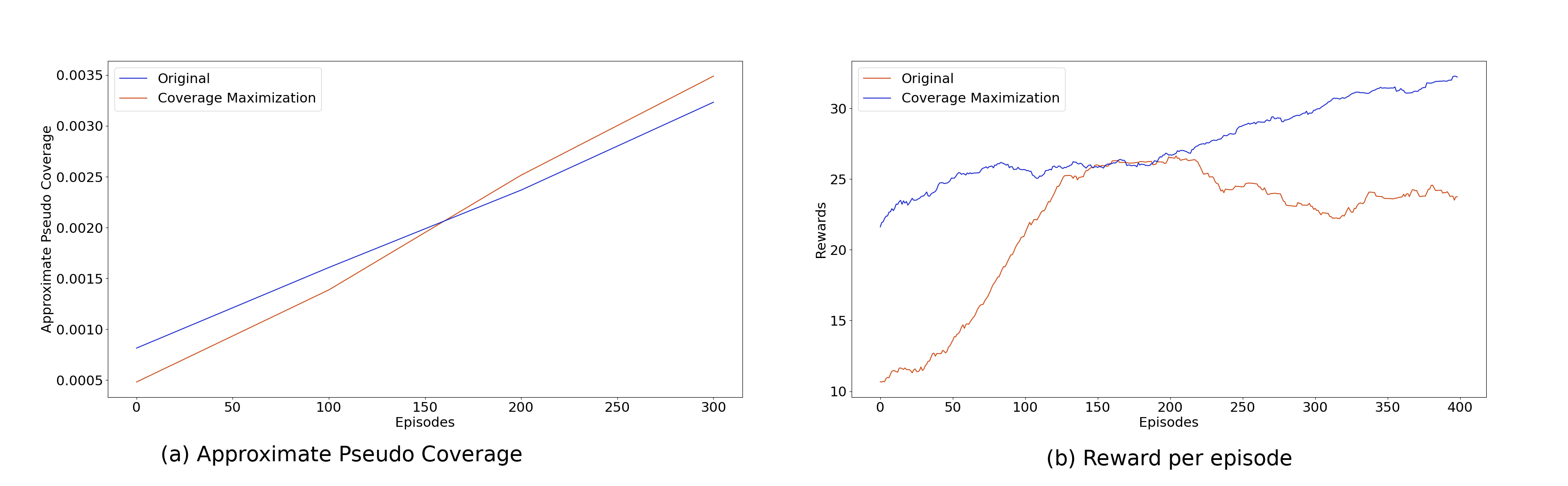}}
         \caption{Comparison of (a) pseudo-coverage metrics and (b) reward per episode for original method and our proposed method with coverage maximization strategy}
        \label{fig:comparison_highway}
\end{figure*}

The state space is $V\times F$ array comprised of $F$ features for $V$ vehicles. The features included are \emph{S = \{presence, x, y, vx, vy\}}. The action space $A$ is discrete with \emph{A = \{left-lane, idle, right-lane, faster, slower\}}. The reward function is composed of a velocity term and a collision term:
\begin{equation}
    R(s,a) = a\times \dfrac{v-v_{min}}{v_{max}-v_{min}} - b \times collision
\end{equation}
where $v, v_{max}$ and $v_{min}$ are the current speed, maximum speed and minimum speed respectively and $a$ and $b$ are coefficients of the velocity and collision penalty terms respectively. 
 
We calculate the coverage metric in a similar fashion to the previous example. Since the initial placement of the vehicles is the most important factor in randomization of the simulation, for constructing the RRT structure, we utilize the longitudinal and lateral positions as abstractions of the entire space for the $V$ vehicles. 

In Figure \ref{fig:comparison_highway}(a) and (b), we provide the coverage metrics and reward for a run of the environment with and without the proposed implementation. The plots show that for a high-dimensional state space DRL problem, our proposed approach is able to increase the coverage while simultaneously maximizing the cumulative reward per episode.

\section{Conclusions}
\label{sec:conclusions}
In this work, we studied the necessity of utilizing coverage as a metric for training of DRL tasks. We proposed a new metric to review the quality of the sampled states for high-dimensional state space. For the two standard tasks, we provided the pseudo-measure of coverage by transforming the state space samples into a 2-D grid and then voting on the occupied grids.  We also proposed using a feedback loop from the training process to the simulation environment and forcing the simulation environment to explore efficiently in the bounds of the state space, both of which led to substantial improvements for the training process.  Thus, we utilized RRT for exploration of unexplored initial states and perturbing high-scoring initial state for exploitation to search for higher reward states. 

For future work, we would like to extend the coverage estimation towards providing convergence guarantees for DRL tasks based on the proposed coverage metric. 

\section*{Acknowledgment}
I would like to thank Dr. Aghapi Mordovanakis for taking the time to edit the manuscript and for stimulating discussions. Also, I would like to mention Dr. Md Tawhid Bin Waez for providing invaluable suggestions during the ideation phase. 

\bibliographystyle{IEEEtran}
\bibliography{biblio}

\begin{thebibliography}{10}
\providecommand{\url}[1]{#1}
\csname url@rmstyle\endcsname
\providecommand{\newblock}{\relax}
\providecommand{\bibinfo}[2]{#2}
\providecommand\BIBentrySTDinterwordspacing{\spaceskip=0pt\relax}
\providecommand\BIBentryALTinterwordstretchfactor{4}
\providecommand\BIBentryALTinterwordspacing{\spaceskip=\fontdimen2\font plus
\BIBentryALTinterwordstretchfactor\fontdimen3\font minus
  \fontdimen4\font\relax}
\providecommand\BIBforeignlanguage[2]{{%
\expandafter\ifx\csname l@#1\endcsname\relax
\typeout{** WARNING: IEEEtran.bst: No hyphenation pattern has been}%
\typeout{** loaded for the language `#1'. Using the pattern for}%
\typeout{** the default language instead.}%
\else
\language=\csname l@#1\endcsname
\fi
#2}}

\bibitem{sutton2018reinforcement}
R.~S. Sutton and A.~G. Barto, \emph{Reinforcement learning: An
  introduction}.\hskip 1em plus 0.5em minus 0.4em\relax MIT press, 2018.

\bibitem{silver2017mastering}
D.~Silver, J.~Schrittwieser, K.~Simonyan, I.~Antonoglou, A.~Huang, A.~Guez,
  T.~Hubert, L.~Baker, M.~Lai, A.~Bolton, \emph{et~al.}, ``Mastering the game
  of go without human knowledge,'' \emph{nature}, vol. 550, no. 7676, pp.
  354--359, 2017.

\bibitem{silver2018general}
D.~Silver, T.~Hubert, J.~Schrittwieser, I.~Antonoglou, M.~Lai, A.~Guez,
  M.~Lanctot, L.~Sifre, D.~Kumaran, T.~Graepel, \emph{et~al.}, ``A general
  reinforcement learning algorithm that masters chess, shogi, and go through
  self-play,'' \emph{Science}, vol. 362, no. 6419, pp. 1140--1144, 2018.

\bibitem{mnih2013playing}
V.~Mnih, K.~Kavukcuoglu, D.~Silver, A.~Graves, I.~Antonoglou, D.~Wierstra, and
  M.~Riedmiller, ``Playing atari with deep reinforcement learning,''
  \emph{arXiv preprint arXiv:1312.5602}, 2013.

\bibitem{levine2016end}
S.~Levine, C.~Finn, T.~Darrell, and P.~Abbeel, ``End-to-end training of deep
  visuomotor policies,'' \emph{The Journal of Machine Learning Research},
  vol.~17, no.~1, pp. 1334--1373, 2016.

\bibitem{hollander_2016}
\BIBentryALTinterwordspacing
Y.~Hollander, ``Machine learning for coverage maximization,'' Sep 2016.
  [Online]. Available:
  \url{https://blog.foretellix.com/2016/09/01/machine-learning-for-coverage-maximization/}
\BIBentrySTDinterwordspacing

\bibitem{araiza2015coverage}
D.~Araiza-Illan, D.~Western, A.~Pipe, and K.~Eder, ``Coverage-driven
  verification—,'' in \emph{Haifa Verification Conference}.\hskip 1em plus
  0.5em minus 0.4em\relax Springer, 2015, pp. 69--84.

\bibitem{byun2020manifold}
T.~Byun and S.~Rayadurgam, ``Manifold for machine learning assurance,''
  \emph{arXiv preprint arXiv:2002.03147}, 2020.

\bibitem{mani2019coverage}
S.~Mani, A.~Sankaran, S.~Tamilselvam, and A.~Sethi, ``Coverage testing of deep
  learning models using dataset characterization,'' \emph{arXiv preprint
  arXiv:1911.07309}, 2019.

\bibitem{bellman1966dynamic}
R.~Bellman, ``Dynamic programming,'' \emph{Science}, vol. 153, no. 3731, pp.
  34--37, 1966.

\bibitem{maaten2008visualizing}
L.~v.~d. Maaten and G.~Hinton, ``Visualizing data using t-sne,'' \emph{Journal
  of machine learning research}, vol.~9, no. Nov, pp. 2579--2605, 2008.

\bibitem{moravec1985high}
H.~Moravec and A.~Elfes, ``High resolution maps from wide angle sonar,'' in
  \emph{Proceedings. 1985 IEEE international conference on robotics and
  automation}, vol.~2.\hskip 1em plus 0.5em minus 0.4em\relax IEEE, 1985, pp.
  116--121.

\bibitem{tavakoli2018prioritizing}
A.~Tavakoli, V.~Levdik, R.~Islam, and P.~Kormushev, ``Prioritizing starting
  states for reinforcement learning,'' \emph{arXiv preprint arXiv:1811.11298},
  2018.

\bibitem{lavalle1998rapidly}
S.~M. LaValle, ``Rapidly-exploring random trees: A new tool for path
  planning,'' 1998.

\bibitem{piziali2008coverage}
A.~Piziali, ``Coverage-driven verification,'' \emph{Functional Verification
  Coverage Measurement and Analysis}, pp. 109--137, 2008.

\bibitem{benjamin1999study}
M.~Benjamin, D.~Geist, A.~Hartman, G.~Mas, R.~Smeets, and Y.~Wolfsthal, ``A
  study in coverage-driven test generation,'' in \emph{Proceedings of the 36th
  annual ACM/IEEE Design Automation Conference}, 1999, pp. 970--975.

\bibitem{lackner2012modeling}
H.~Lackner, H.~Schlingloff, and A.~Berlin, ``Modeling for automated test
  generation-a comparison.'' in \emph{MBEES}.\hskip 1em plus 0.5em minus
  0.4em\relax Citeseer, 2012, pp. 57--70.

\bibitem{lavalle2001rapidly}
S.~M. LaValle and J.~J. Kuffner, ``Rapidly-exploring random trees: Progress and
  prospects,'' \emph{Algorithmic and computational robotics: new directions},
  no.~5, pp. 293--308, 2001.

\bibitem{such2017deep}
F.~P. Such, V.~Madhavan, E.~Conti, J.~Lehman, K.~O. Stanley, and J.~Clune,
  ``Deep neuroevolution: Genetic algorithms are a competitive alternative for
  training deep neural networks for reinforcement learning,'' \emph{arXiv
  preprint arXiv:1712.06567}, 2017.

\bibitem{brockman2016openai}
G.~Brockman, V.~Cheung, L.~Pettersson, J.~Schneider, J.~Schulman, J.~Tang, and
  W.~Zaremba, ``Openai gym,'' \emph{arXiv preprint arXiv:1606.01540}, 2016.

\bibitem{highway-env}
E.~Leurent, ``An environment for autonomous driving decision-making,''
  \url{https://github.com/eleurent/highway-env}, 2018.

\bibitem{mnih2015human}
V.~Mnih, K.~Kavukcuoglu, D.~Silver, A.~A. Rusu, J.~Veness, M.~G. Bellemare,
  A.~Graves, M.~Riedmiller, A.~K. Fidjeland, G.~Ostrovski, \emph{et~al.},
  ``Human-level control through deep reinforcement learning,'' \emph{Nature},
  vol. 518, no. 7540, p. 529, 2015.

\bibitem{polack2017kinematic}
P.~Polack, F.~Altch{\'e}, B.~d'Andr{\'e}a Novel, and A.~de~La~Fortelle, ``The
  kinematic bicycle model: A consistent model for planning feasible
  trajectories for autonomous vehicles?'' in \emph{2017 IEEE Intelligent
  Vehicles Symposium (IV)}.\hskip 1em plus 0.5em minus 0.4em\relax IEEE, 2017,
  pp. 812--818.

\bibitem{treiber2000congested}
M.~Treiber, A.~Hennecke, and D.~Helbing, ``Congested traffic states in
  empirical observations and microscopic simulations,'' \emph{Physical review
  E}, vol.~62, no.~2, p. 1805, 2000.

\bibitem{kesting2007general}
A.~Kesting, M.~Treiber, and D.~Helbing, ``General lane-changing model mobil for
  car-following models,'' \emph{Transportation Research Record}, vol. 1999,
  no.~1, pp. 86--94, 2007.

\end{thebibliography}

\end{document}